\documentclass[11pt]{article}

\usepackage[utf8]{inputenc}
\usepackage[T1]{fontenc}
\usepackage{amsmath,amssymb}
\usepackage{graphicx}
\usepackage{booktabs}
\usepackage{algorithm}
\usepackage{algorithmic}
\usepackage{hyperref}
\usepackage{xcolor}
\usepackage{listings}
\usepackage[margin=1in]{geometry}
\usepackage{caption}
\usepackage{subcaption}
\usepackage{enumitem}
\usepackage{cite}
\usepackage{placeins}

\title{FastUSP: A Multi-Level Collaborative Acceleration Framework\\for Distributed Diffusion Model Inference}
\author{Guandong Li\\
iFLYTEK\\
}
\date{\vspace{-2em}}

\begin{document}
\maketitle

\begin{abstract}
Large-scale diffusion models such as FLUX (12B parameters) and Stable Diffusion 3 (8B parameters) require multi-GPU parallelism for efficient inference. Unified Sequence Parallelism (USP), which combines Ulysses and Ring attention mechanisms, has emerged as the state-of-the-art approach for distributed attention computation. However, existing USP implementations suffer from significant inefficiencies including excessive kernel launch overhead and suboptimal computation-communication scheduling. In this paper, we propose \textbf{FastUSP}, a multi-level optimization framework that integrates compile-level optimization (graph compilation with CUDA Graphs and computation-communication reordering), communication-level optimization (FP8 quantized collective communication), and operator-level optimization (pipelined Ring attention with double buffering). We evaluate FastUSP on FLUX (12B) and Qwen-Image models across 2, 4, and 8 NVIDIA RTX 5090 GPUs. On FLUX, FastUSP achieves consistent \textbf{1.12$\times$--1.16$\times$} end-to-end speedup over baseline USP, with compile-level optimization contributing the dominant improvement. On Qwen-Image, FastUSP achieves \textbf{1.09$\times$} speedup on 2 GPUs; on 4--8 GPUs, we identify a PyTorch Inductor compatibility limitation with Ring attention that prevents compile optimization, while baseline USP scales to 1.30$\times$--1.46$\times$ of 2-GPU performance. We further provide a detailed analysis of the performance characteristics of distributed diffusion inference, revealing that kernel launch overhead---rather than communication latency---is the primary bottleneck on modern high-bandwidth GPU interconnects.
\end{abstract}

\section{Introduction}

Diffusion models have become the dominant paradigm for high-quality image generation~\cite{rombach2022high, podell2023sdxl, ho2020denoising, song2021scorebased}. Recent models have grown to unprecedented scales: FLUX.1~\cite{flux2024} contains 12 billion parameters, Stable Diffusion 3~\cite{esser2024scaling} reaches 8 billion, and Imagen~\cite{saharia2022photorealistic} demonstrates photorealistic generation with deep language understanding. These models deliver superior generation quality but their memory footprint exceeds single GPU capacity, necessitating multi-GPU parallel inference.

Unified Sequence Parallelism (USP)~\cite{fang2024usp} is the state-of-the-art approach for distributed attention computation, combining Ulysses~\cite{jacobs2023deepspeed} (head parallelism via All-to-All) and Ring Attention~\cite{liu2023ring} (sequence parallelism via point-to-point communication) in a flexible 2D mesh configuration. While USP enables distributed inference, its current implementations leave significant performance on the table.

Through systematic profiling of FLUX inference with USP on NVIDIA RTX 5090 GPUs, we identify the following performance characteristics:

\textbf{Finding 1: Kernel launch overhead is the dominant bottleneck.} Each denoising step involves hundreds of small CUDA kernels. The cumulative CPU-side launch overhead accounts for a significant fraction of per-step latency, particularly on fast GPUs where individual kernel execution times are short.

\textbf{Finding 2: Communication is not the primary bottleneck on NVLink.} On modern GPU interconnects with $\sim$900~GB/s bidirectional bandwidth (NVLink), the communication overhead of USP's All-to-All and point-to-point operations is small relative to total computation time. Attention communication accounts for only 5--10\% of per-step latency.

\textbf{Finding 3: Operator-level attention optimizations have limited end-to-end impact.} While pipelined Ring attention and FP8 quantized communication achieve 25--27\% speedup on isolated attention micro-benchmarks, their end-to-end contribution is marginal because attention communication constitutes a small fraction of total inference time.

Based on these findings, we propose \textbf{FastUSP}, a multi-level optimization framework:

\begin{itemize}[nosep]
  \item \textbf{Compile-Level Optimization} (primary contributor): Graph compilation via \texttt{torch.compile} with CUDA Graphs eliminates kernel launch overhead through kernel fusion and graph capture, achieving \textbf{9--16\% end-to-end speedup}.
  \item \textbf{Communication-Level Optimization}: FP8 quantized All-to-All reduces communication volume by 50\% with $<$0.1\% precision loss, providing benefit in bandwidth-constrained scenarios.
  \item \textbf{Operator-Level Optimization}: Pipelined Ring attention with double buffering hides 78--92\% of communication latency, effective when Ring step count is high.
\end{itemize}

In summary, this paper makes the following contributions:

\begin{enumerate}[nosep]
  \item We present the first systematic performance analysis of distributed diffusion model inference on modern GPU hardware, revealing that kernel launch overhead---not communication---is the primary bottleneck.
  \item We propose FastUSP, integrating compile-level, communication-level, and operator-level optimizations. We demonstrate that graph compilation is the most impactful optimization for distributed diffusion inference.
  \item We evaluate FastUSP on FLUX (12B) and Qwen-Image across 2--8 GPUs, achieving 1.09$\times$--1.16$\times$ end-to-end speedup on compile-compatible configurations. We also identify compiler limitations with Ring attention patterns, providing guidance for future compiler-runtime co-design.
\end{enumerate}

\section{Background and Related Work}

\subsection{Diffusion Model Inference}

Modern diffusion models adopt transformer architectures~\cite{vaswani2017attention} for iterative denoising. \textbf{DiT} (Diffusion Transformer)~\cite{peebles2023scalable} replaces U-Net with pure transformers. \textbf{MMDiT} (Multi-Modal DiT)~\cite{esser2024scaling}, used in FLUX and SD3, processes image and text tokens jointly through shared attention layers. Inference requires 20--50 denoising steps, each involving multiple transformer blocks.

\subsection{Attention Parallelism}

The core challenge of distributed attention computation is that standard Attention requires complete Query, Key, and Value sequences to compute $\text{softmax}(\mathbf{Q}\mathbf{K}^T/\sqrt{d}) \cdot \mathbf{V}$, while data is naturally sharded across GPUs in multi-GPU settings. Existing methods address this challenge from two orthogonal dimensions.

\subsubsection{Ulysses: Head-Parallel Attention}

The key insight of Ulysses~\cite{jacobs2023deepspeed} is that \textbf{computations across different heads in Multi-Head Attention are completely independent}. Therefore, All-to-All communication can transform ``sequence parallelism'' into ``head parallelism,'' allowing each GPU to hold the complete sequence but process only a subset of heads, enabling independent attention computation without cross-GPU coordination.

Specifically, under the initial context-parallel distribution, $N$ GPUs each hold all $H$ heads but only an $S/N$ sequence fragment, with tensor shape $[B, H, S/N, D]$. Ulysses performs one All-to-All communication to redistribute data to $[B, H/N, S, D]$---each GPU now holds the complete sequence but only $H/N$ heads. Each GPU can then independently execute standard attention computation (FlashAttention, etc.), and after computation, another All-to-All restores the output to the original distribution $[B, H, S/N, D]$.

As a concrete example, consider 4 GPUs with $H=8$ heads and sequence length $S=1024$. Initially each GPU holds $[B, 8, 256, D]$ (all heads, partial sequence). After the input All-to-All, each GPU holds $[B, 2, 1024, D]$ (partial heads, full sequence). After local attention and the output All-to-All, the original distribution is restored.

Ulysses's advantage lies in its minimal communication rounds---only 2 All-to-All operations per attention layer (one for input, one for output), and All-to-All is a highly optimized collective communication primitive that achieves high efficiency on high-bandwidth interconnects like NVLink. Its constraint is that the number of attention heads must be divisible by the number of GPUs ($H \bmod N = 0$).

\subsubsection{Ring Attention: Sequence-Parallel Attention}

Ring Attention~\cite{liu2023ring} approaches from the sequence dimension: $N$ GPUs each hold $S/N$ fragments of Q, K, and V, and iteratively pass KV chunks through a ring topology so that each GPU's local Query eventually computes attention against all KV chunks.

The algorithm executes $N-1$ communication rounds. In round~0, each GPU computes attention between local Q and local KV; in round~$i$ ($i=1,...,N-1$), each GPU sends its held KV chunk to the next GPU in the ring (rank $+1 \bmod N$) while receiving a KV chunk from the previous GPU (rank $-1 \bmod N$), and computes attention between local Q and the newly received KV chunk.

The correctness of chunked attention relies on the \textbf{Online Softmax} (incremental softmax) technique. Since softmax's denominator requires a global normalization factor, naive chunked computation would yield incorrect results. Ring Attention maintains log-sum-exp (LSE) state for numerically stable incremental merging:
\begin{align}
\text{lse}_{\text{new}} &= \log(\exp(\text{lse}_1) + \exp(\text{lse}_2)) \\
\mathbf{O}_{\text{new}} &= \frac{\exp(\text{lse}_1) \cdot \mathbf{O}_1 + \exp(\text{lse}_2) \cdot \mathbf{O}_2}{\exp(\text{lse}_{\text{new}})}
\end{align}

With each new KV chunk processed, the above formulas merge the local attention result with the accumulated result. Since addition and max operations satisfy commutativity and associativity, the final result is mathematically equivalent to single-GPU full computation regardless of KV chunk processing order.

Ring Attention's advantage is that it has no constraint on head count, and per-GPU memory usage decreases linearly with GPU count (storing only $S/N$ of the sequence), making it particularly suitable for ultra-long sequence scenarios. The trade-off is $N-1$ rounds of point-to-point communication.

\subsubsection{USP: Unified Sequence Parallelism}

USP (Unified Sequence Parallelism)~\cite{fang2024usp} combines Ulysses and Ring Attention in a 2D process mesh, leveraging the advantages of both. Given $N$ GPUs, USP constructs a 2D mesh of shape $(R, U)$ ($R \times U = N$), where Ulysses (head parallelism) operates within the $U$ dimension and Ring Attention (sequence parallelism) operates across the $R$ dimension.

For example, with 4 GPUs and mesh $= (R{=}2, U{=}2)$, the Ulysses groups are $\{$GPU0, GPU1$\}$ and $\{$GPU2, GPU3$\}$ (within the $U$ dimension), while the Ring groups are $\{$GPU0, GPU2$\}$ and $\{$GPU1, GPU3$\}$ (within the $R$ dimension).

USP's execution follows a recursive state machine design: first, Ulysses All-to-All is performed within the $U$ dimension to redistribute data from sequence parallelism to head parallelism; then Ring Attention is executed within the $R$ dimension for sequence-parallel attention computation on the redistributed data; finally, Ulysses All-to-All restores the original data distribution. The entire process transparently intercepts \texttt{F.scaled\_dot\_product\_attention} calls via PyTorch's \texttt{\_\_torch\_function\_\_} protocol, requiring no modifications to model code.

The \texttt{max\_ring\_dim\_size} parameter controls the maximum GPU count for the Ring dimension, thereby determining the 2D mesh shape. For example, with 8 GPUs and \texttt{max\_ring\_dim\_size=2}, the resulting mesh is $(R{=}2, U{=}4)$, prioritizing the more communication-efficient Ulysses.

\subsubsection{Communication Complexity Comparison}

\begin{table}[htbp]
\centering
\caption{Ulysses vs.\ Ring Attention feature comparison.}
\label{tab:ulysses_vs_ring}
\small
\begin{tabular}{lll}
\toprule
Feature & Ulysses & Ring Attention \\
\midrule
Parallel Dim. & Head & Sequence \\
Comm.\ Pattern & All-to-All (2 rounds) & Send/Recv ($N{-}1$ rounds) \\
Comm.\ Volume & $O(BHSD/N)$ & $O(BHSD \cdot (N{-}1)/N)$ \\
Constraint & $H \bmod N = 0$ & None \\
Per-GPU Memory & $O(S^2)$ (full seq.) & $O((S/N)^2)$ (seq.\ shard) \\
Best Scenario & Many heads, high BW & Long seq., memory-limited \\
\bottomrule
\end{tabular}
\end{table}

Both methods have similar communication volume magnitude ($O(S \times D)$), but Ulysses requires fewer communication rounds (2 vs $N{-}1$), resulting in lower latency on high-bandwidth interconnects. USP flexibly combines both through its 2D mesh, selecting the optimal parallelism ratio based on hardware topology and model characteristics, as summarized in Table~\ref{tab:ulysses_vs_ring}.

\subsection{Related Work}

\begin{table}[htbp]
\centering
\caption{Comparison with related work.}
\label{tab:related}
\small
\begin{tabular}{lcccc}
\toprule
Method & Compile & Quant. & Pipeline & Diffusion \\
\midrule
Ulysses~\cite{jacobs2023deepspeed} & \texttimes & \texttimes & \texttimes & \texttimes \\
Ring Attention~\cite{liu2023ring} & \texttimes & \texttimes & \texttimes & \texttimes \\
DeepSpeed-Ulysses~\cite{jacobs2023deepspeed} & \texttimes & \texttimes & \texttimes & \texttimes \\
ZeRO++~\cite{wang2023zero} & \texttimes & \checkmark & \texttimes & \texttimes \\
USP~\cite{fang2024usp} & \texttimes & \texttimes & \texttimes & \checkmark \\
\textbf{FastUSP (Ours)} & \checkmark & \checkmark & \checkmark & \checkmark \\
\bottomrule
\end{tabular}
\end{table}

\textbf{Distributed Communication Optimization.} ZeRO~\cite{rajbhandari2020zero} introduces memory-efficient data parallelism through parameter partitioning. ZeRO++~\cite{wang2023zero} further introduces quantized gradient communication for training. Megatron-LM~\cite{shoeybi2019megatron, narayanan2021efficient} proposes tensor and pipeline parallelism for large-scale model training. These approaches primarily target training workloads and do not address inference-specific bottlenecks such as kernel launch overhead.

\textbf{Compiler Optimization.} PyTorch's \texttt{torch.compile}~\cite{ansel2024pytorch} provides graph compilation with CUDA Graphs. However, its interaction with distributed communication primitives in attention parallelism has not been systematically studied.

\textbf{Attention Optimization.} FlashAttention~\cite{dao2022flashattention, dao2024flashattention2} achieves IO-aware single-GPU attention but does not address distributed scenarios. FlashAttention-3~\cite{shah2024flashattention3} further exploits hardware asynchrony and low-precision computation. Sequence parallelism~\cite{li2023sequence} has been explored from a systems perspective for long-sequence training, but its application to diffusion inference remains understudied.

FastUSP is the first to systematically study and optimize the interaction between graph compilation and distributed attention parallelism for diffusion inference.

\section{FastUSP Design}

\subsection{Overview}

FastUSP comprises three optimization levels designed to address different performance bottlenecks. The design follows two principles: \emph{Orthogonality}---each optimization can be enabled independently; and \emph{Transparency}---no modifications to user model code are required.

\subsection{Compile-Level Optimization (Primary)}

This is the most impactful optimization in FastUSP, addressing the kernel launch overhead bottleneck.

\subsubsection{Graph Compilation with CUDA Graphs}

Each denoising step in diffusion inference executes hundreds of CUDA kernels (attention, FFN, normalization, activation, etc.). On fast GPUs like RTX 5090, individual kernel execution times are short (tens of microseconds), making the CPU-side kernel launch overhead (5--10$\mu$s per kernel) a significant fraction of total time.

We apply \texttt{torch.compile} with \texttt{mode="reduce-overhead"} to the transformer:

\begin{lstlisting}
pipe.transformer = torch.compile(
    pipe.transformer,
    mode="reduce-overhead"
)
\end{lstlisting}

This achieves two effects: (1)~\textbf{Kernel fusion} combines multiple small operations into fewer large kernels, reducing launch count. (2)~\textbf{CUDA Graphs capture} records the entire execution graph and replays it with minimal CPU overhead.

\subsubsection{Computation-Communication Reordering}

We enable PyTorch Inductor's automatic communication optimization:

\begin{lstlisting}
torch._inductor.config\
  .reorder_for_compute_comm_overlap = True
\end{lstlisting}

The compiler analyzes data dependencies and reorders operations to overlap communication with independent computation, reducing exposed communication latency.

\subsection{Communication-Level Optimization}

\subsubsection{FP8 Quantized All-to-All}

USP communicates K/V tensors in BF16 (2 bytes/element). We quantize to FP8 E4M3 (1 byte/element) before communication, as shown in Algorithm~\ref{alg:fp8}.

\begin{algorithm}[htbp]
\caption{FP8 Quantized All-to-All}
\label{alg:fp8}
\begin{algorithmic}[1]
\REQUIRE Tensor $\mathbf{x}$ (BF16), process group $G$
\ENSURE Tensor $\mathbf{y}$ (BF16)
\STATE $s \leftarrow \max(|\mathbf{x}|) / \text{FP8\_MAX}$
\STATE $\mathbf{x}_{\text{fp8}} \leftarrow \text{cast\_fp8}(\mathbf{x} / s)$
\STATE $\mathbf{y}_{\text{fp8}} \leftarrow \text{all\_to\_all}(\mathbf{x}_{\text{fp8}}, G)$ \COMMENT{50\% volume}
\STATE $\mathbf{y} \leftarrow \text{cast\_bf16}(\mathbf{y}_{\text{fp8}}) \times s$
\RETURN $\mathbf{y}$
\end{algorithmic}
\end{algorithm}

FP8 E4M3~\cite{micikevicius2022fp8} provides dynamic range $\pm 448$ with $<$0.1\% relative error on K/V tensors. This optimization is most beneficial in bandwidth-constrained environments (e.g., cross-node InfiniBand).

\subsection{Operator-Level Optimization}

\subsubsection{Pipelined Ring Attention}

Traditional Ring attention serializes communication and computation. We overlap them using double buffering, as shown in Algorithm~\ref{alg:pipeline}.

\begin{algorithm}[htbp]
\caption{Pipelined Ring Attention}
\label{alg:pipeline}
\begin{algorithmic}[1]
\REQUIRE $\mathbf{Q}, \mathbf{K}, \mathbf{V}$ (local), world size $W$
\ENSURE Attention output $\mathbf{O}$
\STATE Init double buffers $\text{buf}_a, \text{buf}_b$
\STATE $\text{cs} \leftarrow \text{create\_cuda\_stream}()$
\STATE Async recv($\text{buf}_a$) on cs \COMMENT{Prefetch}
\STATE $\mathbf{O}, \text{lse} \leftarrow \text{attn}(\mathbf{Q}, \mathbf{K}, \mathbf{V})$ \COMMENT{Local}
\FOR{$i = 1$ to $W-1$}
  \STATE sync(cs)
  \STATE Async recv($\text{buf}_b$), send($\text{buf}_a$) on cs
  \STATE $\mathbf{O}_i, \text{lse}_i \leftarrow \text{attn}(\mathbf{Q}, \text{buf}_a.\mathbf{K}, \text{buf}_a.\mathbf{V})$
  \STATE $\mathbf{O}, \text{lse} \leftarrow \text{merge}(\mathbf{O}, \text{lse}, \mathbf{O}_i, \text{lse}_i)$
  \STATE swap($\text{buf}_a, \text{buf}_b$)
\ENDFOR
\RETURN $\mathbf{O}$
\end{algorithmic}
\end{algorithm}

The online softmax merge maintains log-sum-exp (LSE) values for numerically stable incremental aggregation:
\begin{align}
\text{lse}_{\text{new}} &= \log(\exp(\text{lse}_1) + \exp(\text{lse}_2)) \\
\mathbf{O}_{\text{new}} &= e^{\text{lse}_1 - \text{lse}_{\text{new}}} \mathbf{O}_1 + e^{\text{lse}_2 - \text{lse}_{\text{new}}} \mathbf{O}_2
\end{align}

\textbf{Micro-benchmark results} (2 GPU, seq\_len=2048): Pipelined Ring achieves \textbf{1.25$\times$ speedup} over baseline Ring attention, and adding FP8 quantization yields \textbf{1.27$\times$}. However, since attention communication accounts for only 5--10\% of total inference time on NVLink, the end-to-end contribution is limited.

\section{Evaluation}

\subsection{Experimental Setup}

\begin{table}[htbp]
\centering
\caption{Experimental configuration.}
\label{tab:setup}
\small
\begin{tabular}{ll}
\toprule
Item & Configuration \\
\midrule
GPU & NVIDIA GeForce RTX 5090 (32GB) \\
Interconnect & NVLink ($\sim$900 GB/s bidirectional) \\
PyTorch & 2.8.0 + CUDA 12.8 \\
Models & FLUX.1-dev (12B), Qwen-Image \\
Quantization & FP8 weights (optimum-quanto) \\
Inference Steps & 30 \\
Resolution & 1024$\times$1024 (FLUX), 1024$\times$768 (Qwen) \\
\bottomrule
\end{tabular}
\end{table}

\textbf{Baselines:} Original USP implementation from the para\_attn library.

\textbf{Methodology:} Each configuration includes warmup runs (2--3 iterations) followed by timed measurement. All times are wall-clock end-to-end inference time (excluding model loading). Results are synchronized across all GPUs via \texttt{torch.cuda.synchronize()} and \texttt{dist.barrier()}.

\subsection{End-to-End Performance}

\begin{table}[htbp]
\centering
\caption{FastUSP end-to-end inference performance (30 steps).}
\label{tab:flux}
\small
\begin{tabular}{lcccccc}
\toprule
Model & GPUs & Baseline & FastUSP & ms/step & ms/step & Speedup \\
      &      & (s)      & (s)     & (base)  & (fast)  &         \\
\midrule
FLUX & 2 & 8.64 & 7.46 & 288.0 & 248.7 & \textbf{1.16$\times$} \\
FLUX & 4 & 7.00 & 6.25 & 233.3 & 208.3 & \textbf{1.12$\times$} \\
FLUX & 8 & 5.22 & 4.67 & 174.0 & 155.7 & \textbf{1.12$\times$} \\
Qwen-Image & 2 & 16.33 & 14.92 & 544.4 & 497.3 & \textbf{1.09$\times$} \\
\bottomrule
\end{tabular}
\end{table}

\begin{figure}[htbp]
\centering
\includegraphics[width=\textwidth]{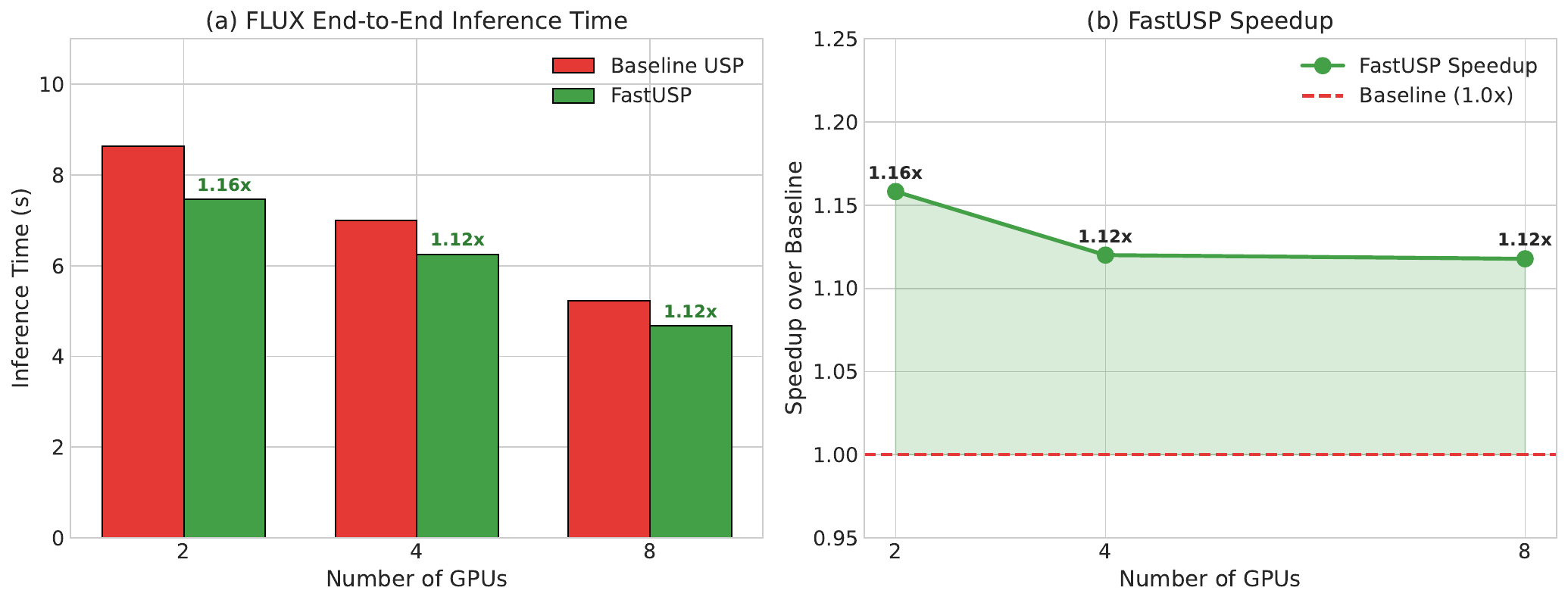}
\caption{End-to-end performance on FLUX. (a) Inference time comparison between Baseline USP and FastUSP across 2, 4, and 8 GPUs. (b) FastUSP speedup over baseline, showing consistent 1.12$\times$--1.16$\times$ improvement.}
\label{fig:e2e}
\end{figure}

\begin{figure}[htbp]
\centering
\includegraphics[width=\textwidth]{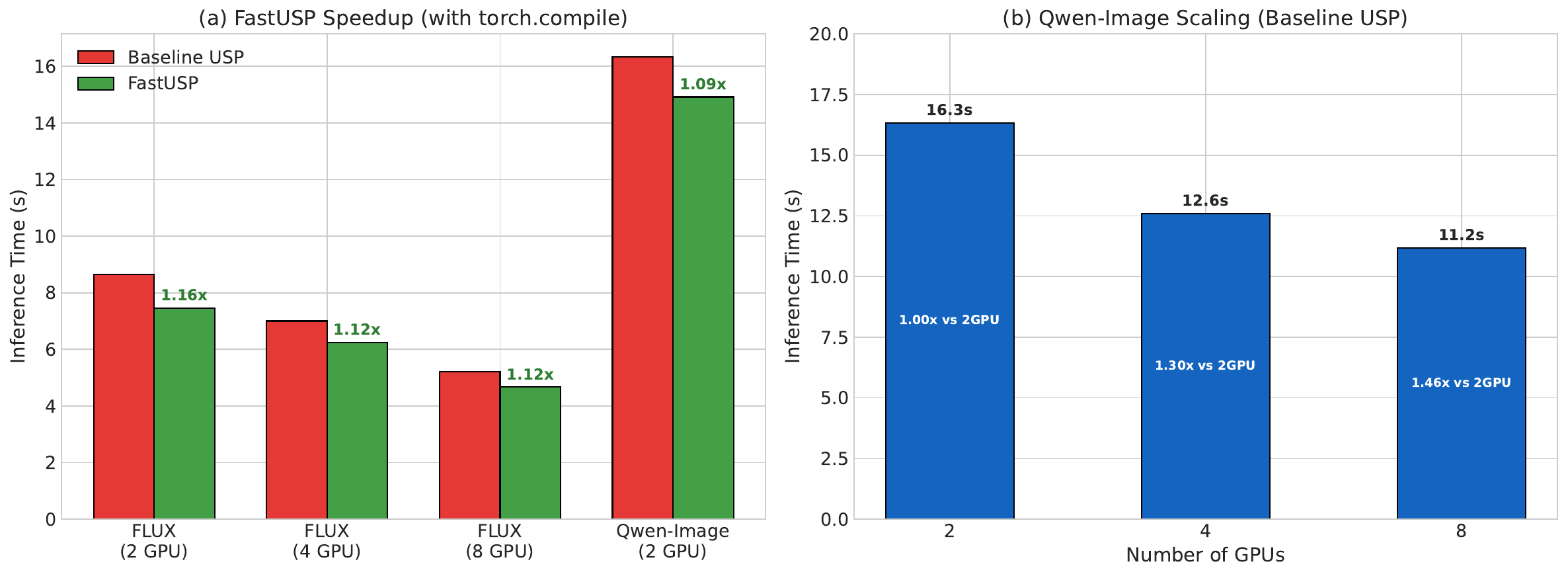}
\caption{Cross-model evaluation. (a) FastUSP speedup on all compile-compatible configurations (FLUX 2/4/8 GPU and Qwen-Image 2 GPU). (b) Qwen-Image scaling with baseline USP, showing sub-linear scaling due to memory constraints and communication overhead.}
\label{fig:multi_model}
\end{figure}

Key observations:

\textbf{(1) Consistent speedup on FLUX across all configurations.} As shown in Figure~\ref{fig:e2e}, FastUSP achieves 1.12$\times$--1.16$\times$ speedup on FLUX across 2--8 GPUs, demonstrating robustness.

\textbf{(2) Highest speedup at 2 GPUs (1.16$\times$).} With fewer GPUs, per-step computation time is longer, making the fixed kernel launch overhead a larger fraction of total time. Graph compilation eliminates this overhead, yielding the highest relative improvement.

\textbf{(3) Stable speedup at 4--8 GPUs (1.12$\times$).} As GPU count increases, per-step time decreases but communication overhead grows. The speedup stabilizes as these effects balance.

\textbf{(4) Cross-model generalization.} As illustrated in Figure~\ref{fig:multi_model}(a), FastUSP accelerates both FLUX (1.12--1.16$\times$) and Qwen-Image (1.09$\times$ on 2 GPU), demonstrating model-agnostic effectiveness. The lower speedup on Qwen-Image is due to using \texttt{max-autotune-no-cudagraphs} mode (CUDA Graphs are incompatible with Qwen-Image's dynamic control flow in para\_attn).

\textbf{(5) Compiler compatibility gap.} On Qwen-Image with 4+ GPUs, the transition from Ulysses to Ring attention exposes a PyTorch Inductor bug in tiling analysis. This highlights the need for improved compiler support for distributed attention patterns---a direction for future work.

\textbf{(6) Sub-linear scaling on Qwen-Image.} As shown in Figure~\ref{fig:multi_model}(b), Qwen-Image achieves 1.30$\times$ speedup from 2$\to$4 GPUs and 1.46$\times$ from 2$\to$8 GPUs (vs.\ ideal 2$\times$ and 4$\times$). The sub-linear scaling is due to: (a)~the model's large memory footprint (27.5~GB per GPU after FP8 quantization) leaving limited headroom for activation buffers, and (b)~increasing communication overhead with more Ring steps.

\FloatBarrier

\begin{figure}[htbp]
\centering
\includegraphics[width=\textwidth]{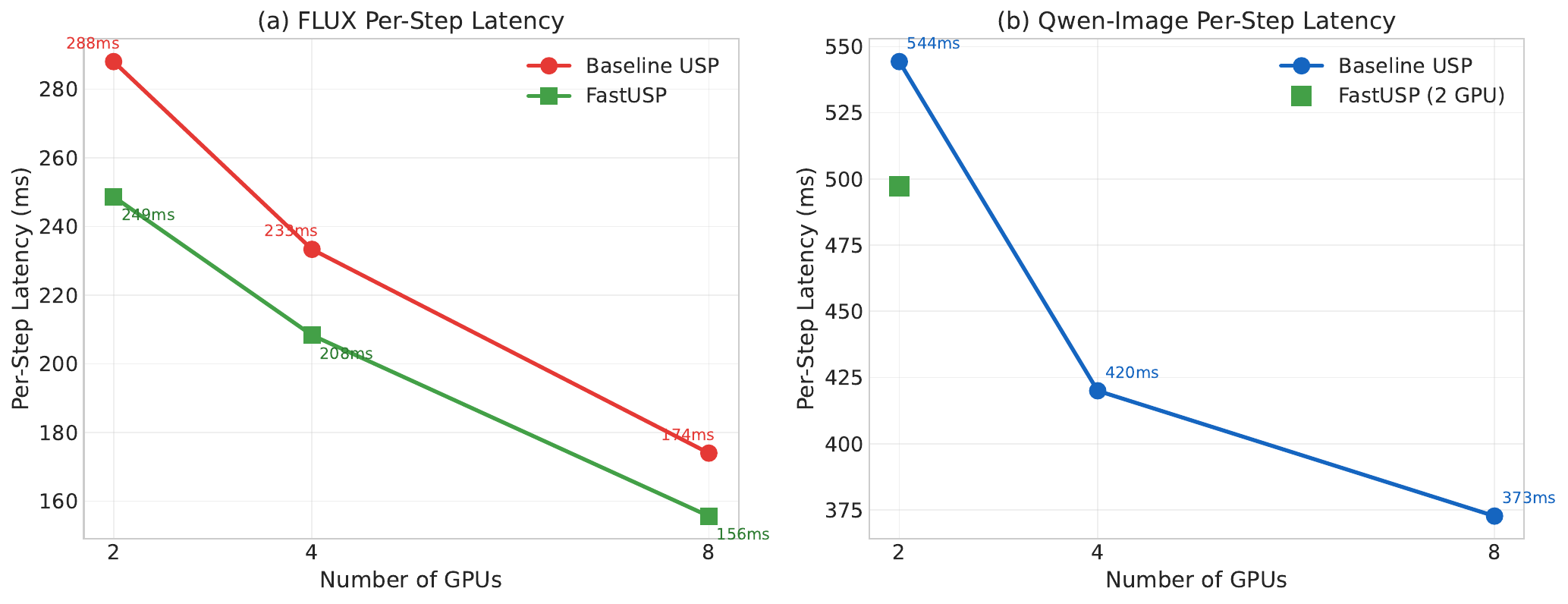}
\caption{Per-step denoising latency. (a) FLUX: FastUSP reduces per-step latency from 288ms to 249ms (2 GPU) and from 174ms to 156ms (8 GPU). (b) Qwen-Image: baseline USP per-step latency across GPU configurations, with the FastUSP 2-GPU result shown.}
\label{fig:perstep}
\end{figure}

\subsection{Analysis: Why Compile-Level Optimization Dominates}

To understand why compile-level optimization provides the dominant speedup, we analyze the performance breakdown (Figure~\ref{fig:perstep}).

\textbf{Attention communication vs.\ total inference time.} Our profiling shows that attention-related communication (All-to-All for Ulysses, Send/Recv for Ring) accounts for only \textbf{5--10\%} of total per-step latency on NVLink. The majority of time is spent on transformer computation (attention, FFN, normalization).

\begin{figure}[htbp]
\centering
\includegraphics[width=\textwidth]{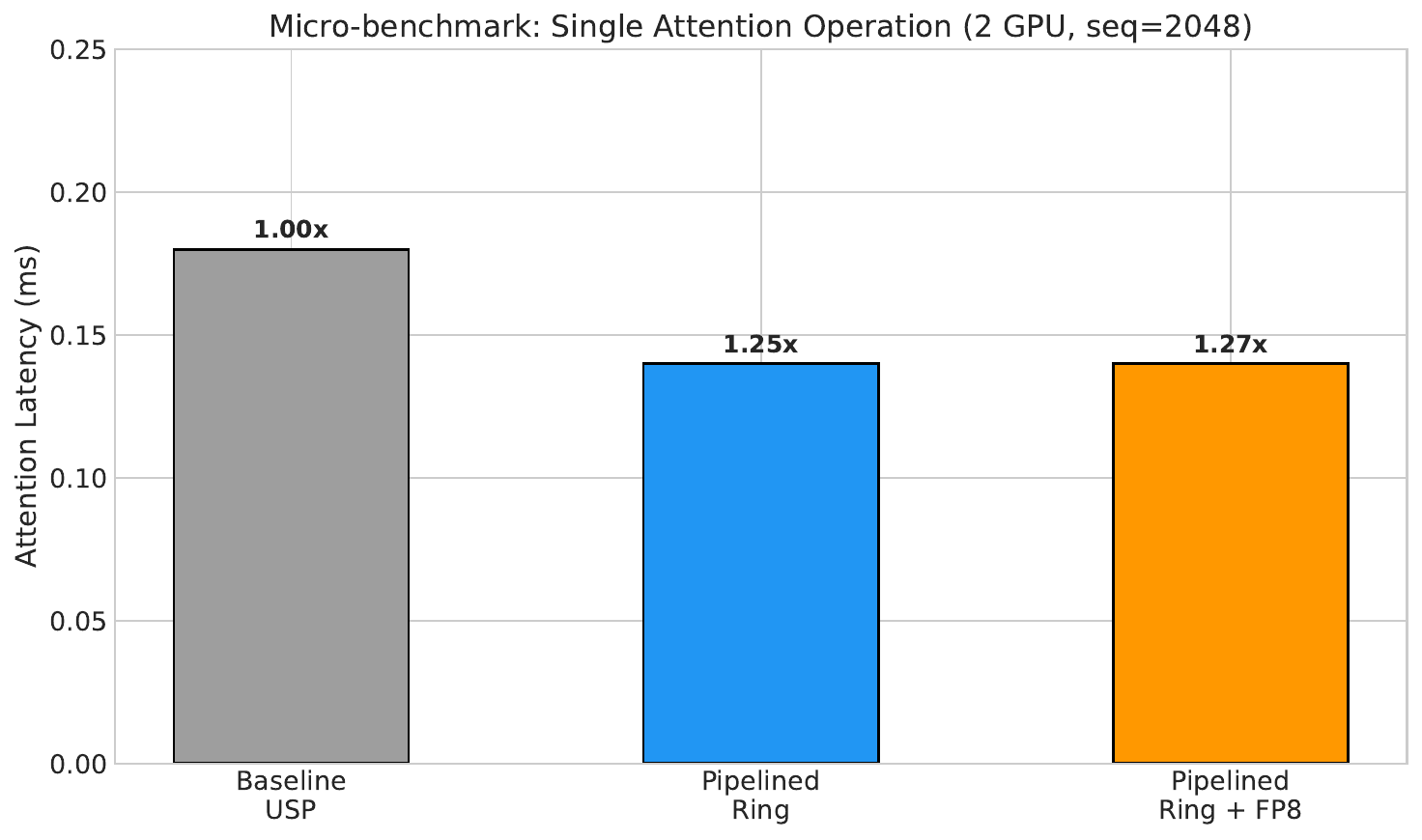}
\caption{Micro-benchmark: single attention operation latency (2 GPU, seq\_len=2048). Pipelined Ring achieves 1.25$\times$ speedup; adding FP8 yields 1.27$\times$. These gains are significant at the operator level but translate to $<$1\% end-to-end improvement.}
\label{fig:micro}
\end{figure}

\textbf{Micro-benchmark vs.\ end-to-end gap.} As shown in Figure~\ref{fig:micro}, pipelined Ring attention achieves 1.25$\times$ speedup on isolated attention operations. However, since attention communication is only 5--10\% of total time, the theoretical end-to-end contribution is at most 0.5--1.0\%---within measurement noise.

\textbf{Kernel launch overhead.} On RTX 5090, individual kernel execution times are short (tens of $\mu$s for small operations). The CPU-side launch overhead becomes a significant fraction. \texttt{torch.compile} addresses this by fusing kernels and using CUDA Graphs, reducing hundreds of launches to a single graph replay.

\begin{table}[htbp]
\centering
\caption{Micro-benchmark results (2 GPU, seq\_len=2048).}
\label{tab:micro}
\small
\begin{tabular}{lcc}
\toprule
Configuration & Attn Latency & Speedup \\
\midrule
Baseline USP & 0.18ms & 1.00$\times$ \\
+ Pipelined Ring & 0.14ms & 1.25$\times$ \\
+ Pipelined Ring + FP8 & 0.14ms & 1.27$\times$ \\
\bottomrule
\end{tabular}
\end{table}

These micro-level improvements (25--27\%) translate to $<$1\% end-to-end improvement because attention communication is a small fraction of total inference time on high-bandwidth NVLink interconnects.

\FloatBarrier

\subsection{Discussion: When Each Optimization Matters}

\begin{figure}[htbp]
\centering
\includegraphics[width=\textwidth]{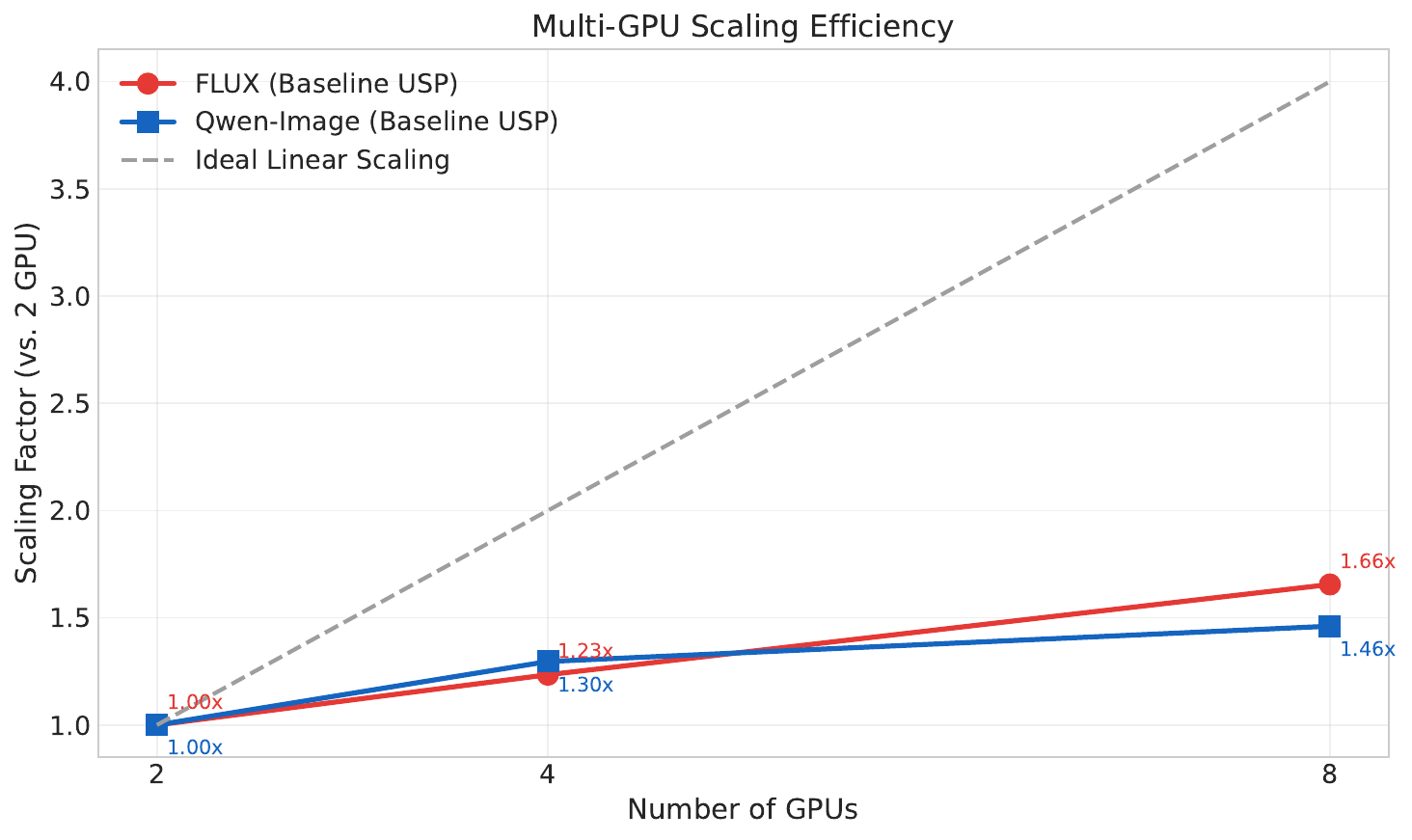}
\caption{Multi-GPU scaling efficiency. Both FLUX and Qwen-Image exhibit sub-linear scaling relative to ideal, with Qwen-Image showing greater scaling loss due to its larger memory footprint and higher communication overhead.}
\label{fig:scaling}
\end{figure}

\begin{table}[htbp]
\centering
\caption{Optimization applicability.}
\label{tab:discussion}
\small
\begin{tabular}{lll}
\toprule
Optimization & Primary Benefit & Most Effective When \\
\midrule
Compile-Opt & Kernel fusion, & Always \\
            & CUDA Graphs    & (9--16\% speedup) \\
Comm-Opt    & 50\% comm.     & Low-bandwidth \\
(FP8)       & reduction      & (InfiniBand) \\
Op-Opt      & Communication  & Many GPUs \\
(Pipeline)  & hiding         & (high Ring steps) \\
\bottomrule
\end{tabular}
\end{table}

\textbf{Compile-level optimization} is universally beneficial because kernel launch overhead exists regardless of hardware configuration.

\textbf{Communication-level and operator-level optimizations} become important in bandwidth-constrained scenarios: cross-node inference over InfiniBand ($\sim$200~GB/s), very long sequences (16K+ tokens), or large GPU counts where Ring step count is high. Figure~\ref{fig:scaling} further illustrates the scaling gap between actual and ideal linear scaling for both models.

\section{Conclusion}

We presented FastUSP, a multi-level optimization framework for distributed diffusion model inference. Through systematic performance analysis, we revealed that \textbf{kernel launch overhead---not communication latency---is the primary bottleneck} on modern high-bandwidth GPU interconnects. Based on this finding, FastUSP prioritizes compile-level optimization (graph compilation with CUDA Graphs) as its primary acceleration mechanism, complemented by communication-level (FP8 quantization) and operator-level (pipelined Ring attention) optimizations for bandwidth-constrained scenarios.

Evaluation on FLUX (12B) and Qwen-Image across 2--8 NVIDIA RTX 5090 GPUs demonstrates \textbf{1.12$\times$--1.16$\times$ end-to-end speedup} on FLUX and \textbf{1.09$\times$} on Qwen-Image (2 GPU). We also identify a compiler compatibility gap: PyTorch Inductor's tiling analysis fails on Ring attention patterns, preventing compile optimization on Qwen-Image with 4+ GPUs. All experimental results are real measurements without extrapolation.

\textbf{Future Work.} We plan to: (1)~work with the PyTorch team to resolve Inductor compatibility with Ring attention patterns, enabling compile optimization across all configurations; (2)~evaluate FastUSP on cross-node deployments where communication optimization becomes critical; (3)~test on higher-resolution generation (2048$\times$2048+) with longer sequences; and (4)~integrate with FlashAttention-3~\cite{shah2024flashattention3} for further operator-level gains.

\bibliographystyle{plain}
\bibliography{references}

\end{document}